\documentclass[conference]{IEEEtran}
\usepackage{amsmath,amssymb,amsfonts}

\usepackage{algpseudocode}
\usepackage{graphicx}
\usepackage{textcomp}
\usepackage{algorithm,setspace}
\usepackage{mathtools}
\usepackage{subcaption}
\usepackage[colorlinks=true, allcolors=black]{hyperref}
\usepackage{url}
\usepackage[flushleft]{threeparttable}
\usepackage{xcolor}

\usepackage{fancyhdr}

\def\BibTeX{{\rm B\kern-.05em{\sc i\kern-.025em b}\kern-.08em
    T\kern-.1667em\lower.7ex\hbox{E}\kern-.125emX}}
    
\begin{document}

\title{Wafer2Spike: Spiking Neural Network for Wafer Map Pattern Classification}

\author{\IEEEauthorblockN{Abhishek Mishra, Suman Kumar, Anush  Lingamoorthy, Anup Das, and Nagarajan Kandasamy}
\IEEEauthorblockA{Electrical and Computer Engineering Department \\
Drexel University \\
Philadelphia, Pennsylvania 19104
}}
% \and
% \IEEEauthorblockN{Name2}
% \IEEEauthorblockA{Twentieth Century Fox\\
% Springfield, USA\\
% Email: homer@ilikeitc2021.com}
% \and
% \IEEEauthorblockN{Name3\\ and Name4}
% \IEEEauthorblockA{Starfleet Academy\\
% San Francisco, California 96678--2391\\
% Telephone: (800) xxx--xxxx\\
% Fax: (888) yyy--yyyy}}

\maketitle

\begin{abstract}
In integrated circuit design, the analysis of wafer map patterns is critical to improve yield and detect manufacturing issues. We develop Wafer2Spike, an architecture for wafer map pattern classification using a spiking neural network (SNN), and demonstrate that a well-trained SNN achieves superior performance compared to deep neural network-based solutions. Wafer2Spike achieves an average classification accuracy of 98\% on the WM-811k wafer benchmark dataset. It is also superior to existing approaches for classifying defect patterns that are underrepresented in the original dataset. Wafer2Spike achieves this improved precision with great computational efficiency. 
\end{abstract}


\section{Introduction}
%As integrated circuit (IC) technology scales, fatal defects and process variations have emerged as one of the key factors limiting yield~\cite{moore2020international}. These process variations arise from various systematic issues; for example, misalignment between contacts and gates due to a malfunctioning phototool or stress-induced cracking during the chemical-mechanical polishing phase. A critical step toward improving the yield involves identifying the primary causes of loss throughout the IC design cycle. In this context, 

Wafer pattern classification involves analyzing the patterns on silicon wafers to identify defects and irregularities that could affect the performance and reliability of the final product. By accurately classifying these patterns, manufacturers can quickly address process issues, reduce waste, and improve yield. Deep neural networks (DNNs) are very effective in classifying wafer map patterns because of their capacity to learn directly from raw data and their strong generalization ability. However, they require significant amounts of computing resources for training and inference. A promising alternative is \emph{spiking neural networks} (\emph{SNNs}) inspired by the operational principles of biological neurons in which neurons communicate with each other by sending short impulses, called spikes, through synapses. This form of brain-inspired computing holds significant promise for various spatial and temporal pattern recognition tasks. SNNs process information through discrete spikes, which are inherently well suited for handling high-dimensional spatial patterns. The ability to process inputs sparsely also reduces noise, allowing the network to focus on subtle details and variability within these patterns. Furthermore, when coupled with neuromorphic hardware, SNNs can deliver improved accuracy with great computational efficiency.  

This paper demonstrates the application of SNN to wafer map pattern classification and shows that a well-trained SNN can attain superior accuracy compared to current DNN-based solutions. Wafer2Spike is a domain-specific SNN designed to recognize wafer map patterns with high precision. The design addresses challenges posed by SNNs --- extreme sparsity, non-differentiable operators and reliance on approximate gradients, and single-bit activations --- through appropriate input encoding, network architecture, and tailored training strategies.

The performance of Wafer2Spike in terms of classification accuracy is evaluated using the WM-811k wafer benchmark dataset~\cite{wu2014wafer}. The results show that Wafer2Spike outperforms state-of-the-art DNN-based approaches with the proposed data augmentation method, achieving an average classification accuracy of 98\%. It also achieves superior accuracy on important underrepresented patterns compared to existing methods. Wafer2Spike is also computationally efficient: using energy consumption as a proxy for efficiency, we show significant savings, up to 22 times, over DNN models.
 
%The paper is organized as follows. Section~\ref{sec:background} provides background on the wafer pattern classification problem and describes the basic concepts underlying SNNs, including training approaches for these networks. Section~\ref{sec:arch} develops the Wafer2Spike architecture. Section~\ref{sec:evaluation} describes the experimental setup and discusses the key results. Section~\ref{sec:related_work} discusses related work, and we conclude the paper in Section~\ref{sec:conclusion}.  

\section{Background}\label{sec:background}
We describe the wafer map pattern classification problem and familiarize the reader with basic concepts related to spiking neural networks.

\subsection{Wafer Map Pattern Classification}
\begin{figure}[t!]
\centering
\includegraphics[width=0.85\columnwidth]{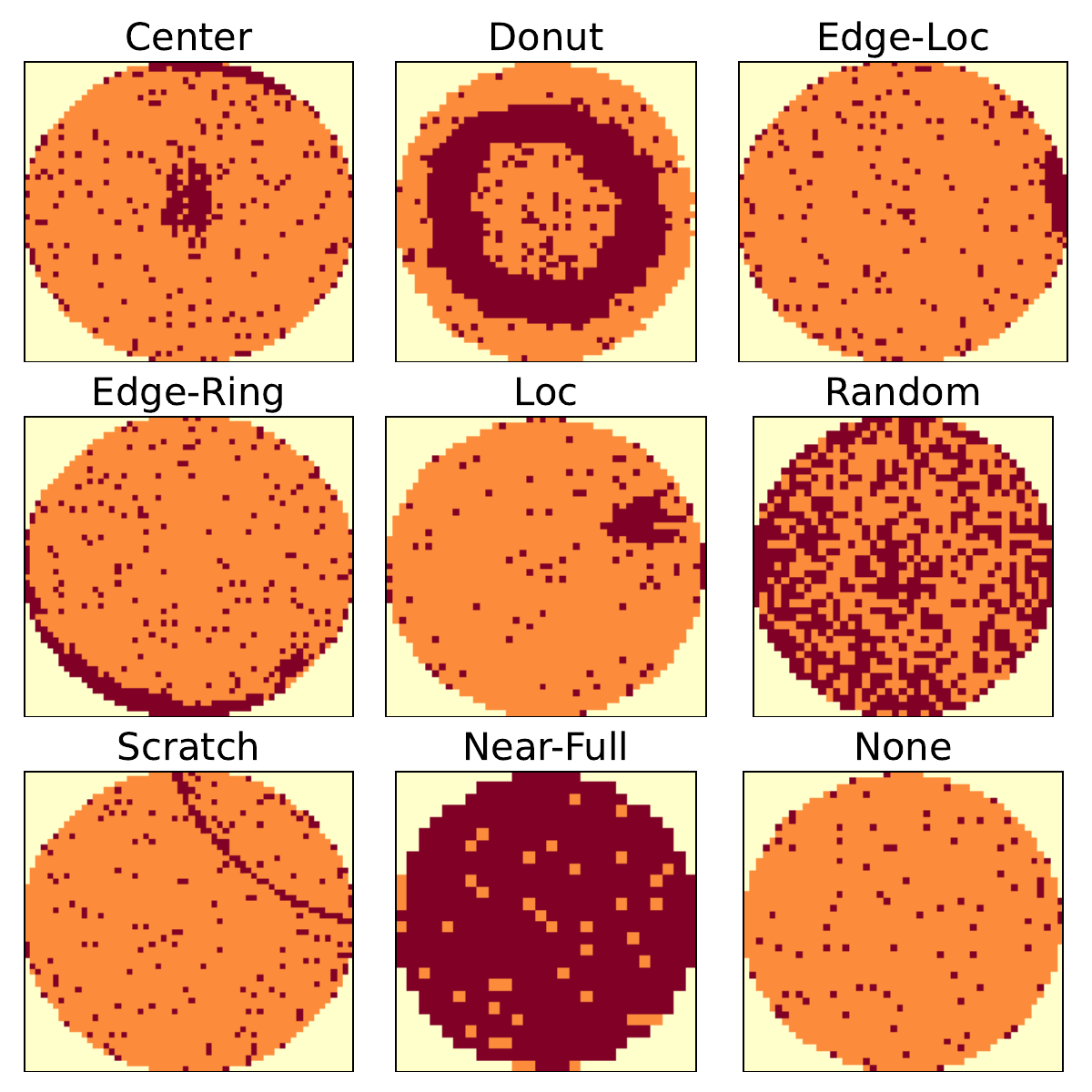}
\caption{Examples of different wafer map patterns present in the WM-811k dataset (best viewed in color).} 
\vspace{-12pt}
\label{fig:wafer_patterns}
\end{figure}

Consider wafer datasets $\mathbf{D_{wf}} = \{D_{wf^1}, D_{wf^2}, \ldots, D_{wf^n}\}$ and classes $\mathbf{C}= \{C_1, C_2, \cdots,C_n\}$ where every instance within $D_{wf^i} \in \mathbf{D_{wf}}$ is labeled with the same class $C_i \in \mathbf{C}$. Figure~\ref{fig:wafer_patterns} shows the nine wafer map patterns or classes present in the WM-811k dataset: Center, Donut, Edge-Location, Edge-Ring, Random, Location, Near-Full, Scratch, and No-Pattern. Orange pixels signify a good die that has successfully passed all wafer tests, red pixels indicate a bad die that has failed some test, and yellow pixels represent regions outside the wafer. The training set $\mathbf{T_{wf}} = \{(d_{wf^i},C_i)\}_1^N$, contains $N$ examples; $d_{wf^i}$ is an example from the wafer dataset associated with the label $C_i$. Our objective is to train a classifier, Wafer2Spike($d_{wf}$), using the training set $\mathbf{T_{wf}}$ and evaluate its performance on a testing set $\mathbf{Te_{wf}}$.

\subsection{Spiking Neural Network Architecture}
An SNN is modeled as a graph in which the vertices represent spiking neurons and the directed edges represent the synapses between neurons. The weight along each edge represents the strength of the synaptic connection between two neurons. We use a first-order current-based leaky integrate-and-fire (LIF) neuron as our fundamental unit within the SNN~\cite{izhikevich2004model}. The neurons accumulate inputs through a weighted sum, which influences the $j^{\mathrm{th}}$ neuron's membrane potential ${V_t}^{L_j}$ in the $L^{\mathrm{th}}$ layer of the network at time $t$. When the combined signal raises the neuron's membrane potential above a threshold $V_{\mathrm{thr}}$, the neuron fires a spike ${Spk}_t$ and resets its potential ${V_t}^{L_j}$ to a baseline value $V_{\mathrm{reset}}$.
% \begin{equation} \label{eq:Eq1}
% {{Spk}_t}^{L_j} = 
% \begin{cases} 
% 1$ \& ${V_t}^{L_j} = V_{reset}, & \text{if } {V_t}^{L_j} > V_{thr}; \\
% 0, & \text{if } {V_t}^{L_j} \leq V_{thr}
% \end{cases}
% \end{equation}

The electrical behavior of a neuron’s membrane potential can be modeled as a resistor-capacitor (RC) circuit that captures the membrane's capacity to charge and discharge in response to electrical impulses, mimicking the neuron’s response to synaptic inputs. By solving the underlying differential equations for the RC circuit, the dynamics of the membrane potential over time can be approximated as 
\begin{align} \label{eq:Eq2}
{Isc}^L_t &= W_{scd}^L \cdot {Isc}^L_{t-1} + f_{v}({Spk}^{L-1}_t) \; \mathrm{and} \\
{V_t}^L &= W_{vd}^L \cdot V_{t-1}^L + {Isc}^L_t. \nonumber
\end{align} 
The equations describe the mechanism by which input spikes are translated to synaptic current ${Isc}^L_t$, which in turn contributes to the membrane voltage ${V_t}^L$ for the $L^{\mathrm{th}}$ layer at time $t$.
Here, ${W_{scd}^L} \in R^+$ represents the matrix of synaptic current decay factors and is applied to each LIF neuron within the $L^{\mathrm{th}}$ layer. The function $f_{v}$ generates the postsynaptic potential as specified by the $L^{\mathrm{th}}$ layer, ${Spk}^{L-1}_t$ denotes the occurrence of spikes from neurons from the previous layer, and $W_{vd}^L$ is the voltage decay factor for each neuron in the $L^{\mathrm{th}}$ layer. Both $W_{scd}^L$ and $W_{vd}^L$ are treated as training parameters in our architecture. Gradient descent is used to optimize these parameters, following the work by Wu et al. on task-dependent hyperparameter optimization~\cite{wu2018spatio}.

\begin{figure*}[t!]
\centering
\includegraphics[width=0.99\textwidth]{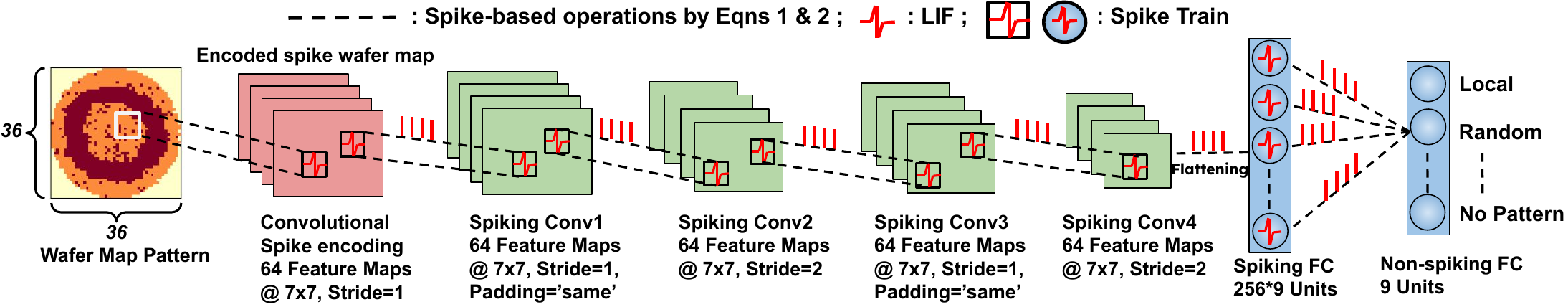}
\caption{The Wafer2Spike architecture comprising the convolutional spike encoding layer, spike-based convolutional layers, a fully-connected spiking layer, and the non-spiking output layer.} 
\label{fig:Wafer2Spike}
%\vspace{-12pt}
\end{figure*}

\begin{figure*}[t!]
\centering
\includegraphics[width=0.9\textwidth]{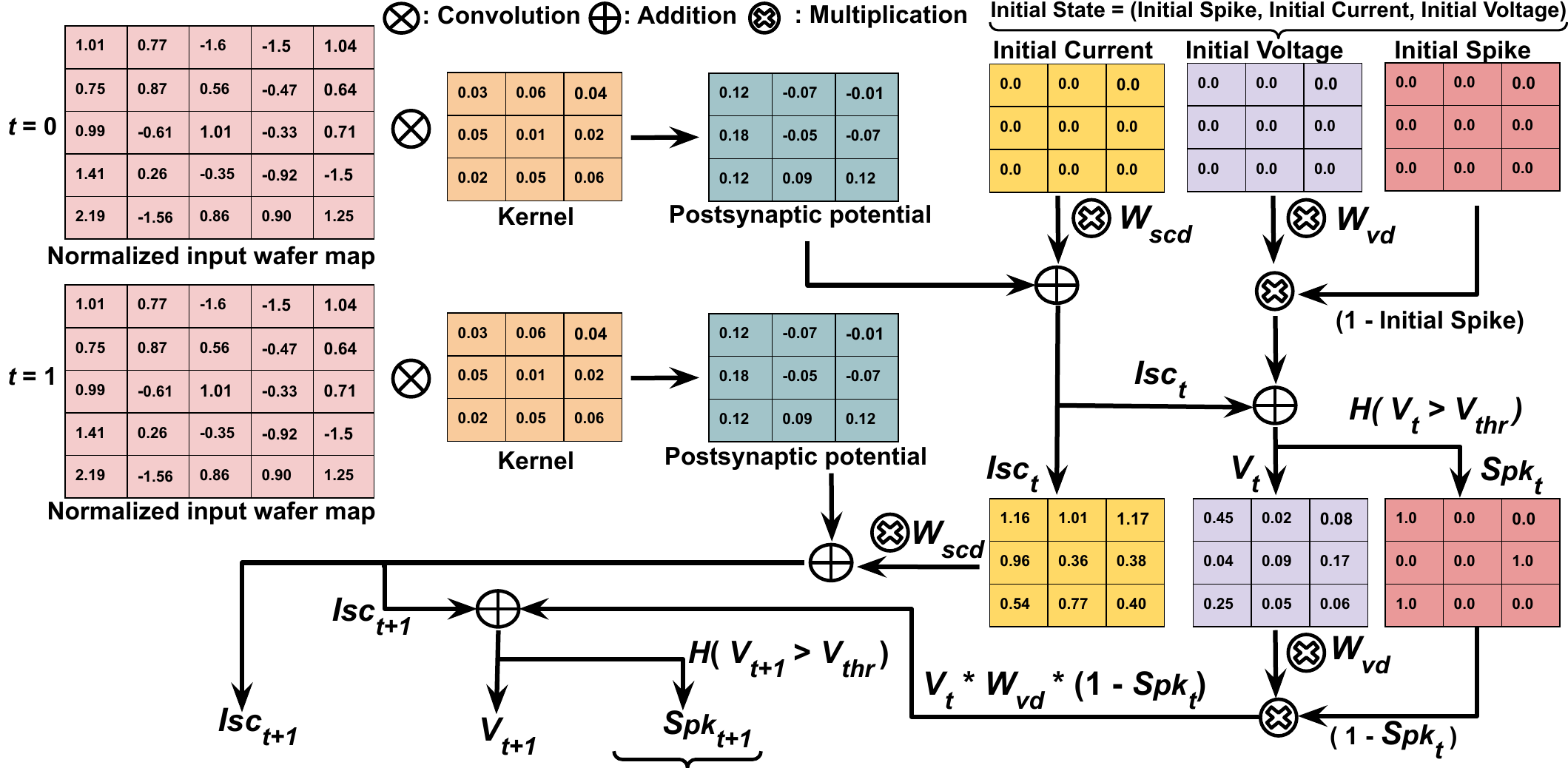}
\caption{Convolutional spike encoding operation for the first two time steps.} 
\vspace{-12pt}
\label{fig:Conv_Spk_encd}
\end{figure*}

\subsection{Training of Spiking Neural Networks}
 \emph{Direct training approaches} take advantage of the unique properties of SNNs by incorporating spike timing information during the training process and can be implemented in both supervised and unsupervised learning paradigms, using the precise timing of spikes to encode and process information. Due to the inherent non-differentiability of spiking neurons, surrogate gradients are typically used to guide backpropagation. This form of training is inherently more aligned with the natural operation of SNNs. \emph{DNN-to-SNN approaches} train a traditional DNN and then converts it to an SNN. The conversion process ensures that the SNN retains the same functional mapping as its non-spiking counterpart, allowing for the seamless transition of learned behaviors and patterns into the spiking domain~\cite{bu2023optimal}. These methods address the challenge posed to direct training methods due to the non-differentiability of spikes.

The DNN-to-SNN conversion process is an approximation which imposes a significant burden on the layer-to-layer bandwidth capacity due to binarized spike activations. This often results in a decrease in the accuracy of the resulting SNNs. It also extends the simulation time steps needed to accurately convert high-precision activations into spikes, resulting in increased latency during inference. Therefore, our work uses a spike-based direct supervised training method along with a backpropagation technique specifically tailored for spikes.

\section{The Wafer2Spike Architecture}\label{sec:arch}
The Wafer2Spike architecture shown in Fig.~\ref{fig:Wafer2Spike} consists of a convolutional spike encoding layer to generate spikes from the input wafer map pattern, followed by multiple spike-based convolutional layers to extract spatio-temporal features, terminated by a fully connected spiking and output layer. 
\vspace{6pt}

\noindent \emph{Convolutional Spike Encoding Layer:} Pixel intensity values are used within the wafer map to generate spikes with the appropriate frequency and inter-spike interval. The encoding operation transforms the static input data into dynamic, temporal spike patterns. Figure~\ref{fig:Conv_Spk_encd} shows an example of this operation over two time steps. The wafer map is resized to a dimension of $(1, 36, 36)$ and processed by a convolution operation to extract features using a $7 \times 7$ kernel which moves across the map with a stride length of one. No padding is applied during the operation. Consequently, 64 distinct feature maps are generated. The value of each point ${P^{c_i}}(x, y)$ within these feature maps is calculated as 
\begin{equation}\label{eq:Eq4}
\left (\sum_{d} \sum_{j} \sum_{k} IW_{d}(x + j, y + k) 
\times {W_{d}}^{c_i}(j, k) \right) + b^{c_i},
\end{equation}
where $d$ iterates over the depth dimensions of both the input volume $IW$ and the learnable kernel ${W_{d}}^{c_i}$. The coordinates $x$ and $y$ represent a specific point within the input wafer map, while ${W_{d}}^{c_i}(j,k)$ refers to the kernel element at the position $(j,k)$ across the depth dimension $d$ for the ${c_i}^{\mathrm{th}}$ feature map. The term $b^{c_i}$ is the bias associated with the ${c_i}^{\mathrm{th}}$ feature map. 

Convolutions output $P^{c_i} \in (batch\_size, 64, 30, 30)$ which is used as the potential function $f_{v}$ for input encoding purposes to produce postsynaptic potential values. Subsequently, each value (neuron) within the feature maps is processed through the LIF operation cycle described in section \ref{sec:background}, converting the continuous intensity values in the wafer maps into binary spike events $P^{c_i}(Spk_t)(x, y)$ while preserving their spatial relationships. The encoding process itself is trainable and thus does not require manual parameter tuning. \vspace{6pt}

%In summary, spikes generated by the convolutional spike encoding layer indicate the presence or absence of spike events at each feature map location, seamlessly transitioning from encoding to further processing within the SNN. 

%By streamlining the encoding process, our method enhances adaptability and aligns with the dynamic requirements of neural data processing.

% Since the wafer map is a grayscale image, the depth is set to $d = 1$.

\noindent \emph{Spiking-Based Convolutional Layer:} 
This convolutional layer mimics a LIF's neuronal behavior and processes spikes through convolutions adjusted through synaptic weights, allowing it to extract higher-level features by capturing complex spatial dependencies. The convolution operation is given by 
\begin{align}\label{eq:Eq5}
&f_{conv}^{c_i}(P^{c_i}(Spk_t))(x, y) = \\ \nonumber
&\left( \sum_{d} \sum_{j} \sum_{k} \underset{d}{P^{c_i}}(Spk_t)(x + j, y + k) \cdot {W^L_{d}}^{c_i}(j, k) \right) + {b^L}^{c_i}
\end{align}
where $f_{conv}^{c_i}$ calculates the postsynaptic potential at location $(x,y)$ within the ${c_i}^{\mathrm{th}}$ feature map of the convolutional layer in the $L^{\mathrm{th}}$ layer. This layer uses learnable parameters, kernel weights ${W^L_{d}}^{c_i}$ and biases ${b^L}^{c_i}$, to fine-tune the network response in the $L^{\mathrm{th}}$ layer. Our SNN architecture allows for up to four subsequent spiking-based convolutional layers for more intricate spatial feature extraction throughout the network. LIF operations are implemented within each spiking-based convolutional layer to produce spike events. \vspace{6pt}

\noindent \emph{Spiking-Based Fully Connected Layer:} 
The fully connected layer within Wafer2Spike integrates the spatial characteristics extracted by the preceding spiking-based convolutional layers using a linear transformation. This synthesis enables the network to identify complex patterns throughout its training, improving its ability to approximate the target function. The linear transformation is calculated as
\begin{equation}\label{eq:Eq6}
f_{fcv}(Spk^{L-1}_t) = W^L_{fc} \cdot Spk^{L-1}_t + b^L_{fc},
\end{equation}
where $f_{fcv}$ provides the postsynaptic potential for the fully connected layer $L$, $Spk^{L-1}_t$ denotes the spike inputs received from layer $L-1$ (the last spiking-based convolutional layer). This layer also uses learnable parameters to optimize its behavior---specifically, the synaptic weights $W^L_{fc}$ associated with each neuron and the bias vector $b^L_{fc}$. \vspace{6pt}

\noindent \emph{Non-spiking-based fully connected layer:}The non-spiking fully connected layer operates analogously to the output layer within a DNN. It functions as a decoder in the case of the SNN to translate the spike-based representation into a more understandable output: a vector of class probabilities for each time step. This translation is performed by the following equations.  
\begin{align}\label{eq:Eq7}
 & prob_t = W^L_{nspk} \cdot Spk^{L-1}_t + b^L_{nspk} \; \mathrm{and} \nonumber \\ 
 & Prob_{class} = \sum_{t} W_t \cdot prob_t. 
\end{align}
Here, $W^L_{nspk}$ and $b^L_{nspk}$ are the learnable parameters of the non-spiking fully connected layer, and $Spk^{L-1}_t$ denotes the incoming spikes from the preceding spiking-based fully connected layer. Upon receiving these spikes, \eqref{eq:Eq7} calculates the time-specific vector of class probabilities $prob_t$ using time-dependent weight parameters $W_t$. Subsequently, the final class probabilities $Prob_{class}$ are calculated as a weighted sum of these probabilities across all time steps. \vspace{6pt}

%In summary, the non-spiking-based fully connected layer consolidates and interprets the features learned by previous layers, acting as a decoder that maps complex feature representations that are originally in the form of spikes to understandable output predictions.

\noindent \emph{SNN Training Procedure:}
Wafer2Spike is trained using a cross-entropy loss function $L_{ce}$ to classify $C$ distinct wafer map patterns.
\begin{equation}\label{eq:LCE}
    L_{ce}(y, Prob_{class}) = -\frac{1}{N} \sum_{i=1}^{N} \sum_{c=1}^{C} y_{ic} \cdot \log(Prob_{class_{ic}}).
\end{equation}
Here, $y_{ic}$ is an indicator of the correct classification of the class label $c$ for observation $i$, $Prob_{class_{ic}}$is the predicted probability that observation $i$ is of class $c$, and $N$ is the total number of wafer pattern samples for training. As the LIF model demonstrates, spike signals propagate not just through the spatial domain from layer to layer but also influence neuronal states across the temporal domain. Consequently, gradient-based training methods must take into account derivatives in both spatial and temporal dimensions. Therefore, we have incorporated the weights of synaptic current decay $W_{scd}$ and voltage decay $W_{vd}$ of the LIF neuron along with other SNN parameters into a spatio-temporal backpropagation method to minimize the average cross-entropy loss $L_{ce}$ over all the training samples. We follow an approach developed by Wu et al.~\cite{wu2018spatio} that uses a surrogate gradient to guide backpropagation.

\section{Performance Evaluation}\label{sec:evaluation}
This section evaluates Wafer2Spike in terms of accuracy and computational efficiency against other competing approaches.

\subsection{The WM-811K Dataset and Data Augmentation}
The WM-811K dataset aggregates 811,457 wafer maps sourced from 46,293 individual lots, featuring nine unique defect patterns: Center, Donut, Edge-Location, Edge-Ring, Random, Location, Near-Full, Scratch, and No-Pattern, which were previously shown in Fig.~\ref{fig:wafer_patterns}. The wafer maps vary in size and contain three pixel classification levels: '2' indicates defective dies, '1' represents operational dies, and '0' corresponds
to the absence of a die. To standardize the input, all wafer maps are resized to a consistent resolution of $36 \times 36$ pixels. In the visualization provided in Fig.~\ref{fig:wafer_patterns}, the operational dies are depicted in orange, marking the units that meet the test criteria; defective dies are marked in red, indicating those that fail the tests; and the yellow pixels mark the non-die regions on the wafer. The WM-811k dataset comprises 172,950 wafer maps labeled by domain experts. Thus, for a fair comparison with competing approaches~\cite{wu2014wafer, alawieh2020wafer, genssler2021brain, tsai2020light, liao2022wafer}, this study uses only labeled data and categorizes the data into distinct groups, each corresponding to different training-to-test ratios.

During IC production, wafer maps are generated without displaying any distinct patterns much more frequently than those that exhibit specific patterns. This leads to a bias towards predicting the more common No-Pattern category of wafers and negatively impacts the detection of critical but infrequent patterns. Our \emph{data augmentation technique} uses bijective functions to produce a specified amount of analogous data from underrepresented patterns while maintaining the intrinsic characteristics of the original data. For each minority class, we randomly select certain images as templates and then apply a series of geometric transformations to these to infuse variation, thereby creating new, diverse patterns.

\begin{table*}[t!]
\begin{threeparttable}
\small
\centering
\fontsize{6.5}{12}\selectfont
\caption{Accuracy achieved by Wafer2Spike compared to other approaches.}%

\begingroup
\setlength{\tabcolsep}{6pt} % Default value: 6pt
\renewcommand{\arraystretch}{.85}
  \begin{tabular}{|ccccccccccccc|}
    \hline
    Reference & Model & Split Ratio & Avg. Accuracy & No Pattern & Center & Donut & Edge-Loc & Edge-Ring & Local & Random & Scratch & Near-Full \\
    \hline
    \cite{alawieh2020wafer} & CNN & 8:2 & 94\% & 98\% & 96\% & 73\% & 70\% & 96\% & 64\% & 57\% & 29\% & 40\%\\
    \cite{alawieh2020wafer} & SVM \cite{wu2014wafer}$^H$ & 8:2 & 91\% & \textbf{100\%} & 79\% & 36\% & 50\% & 96\% & 2\% & 0\% & 38\% & 80\%\\
    \cite{genssler2021brain} & MC32+MLP & 8:2 & 96\% & 98\% & 96\% & 89\% & 87\% & 98\% & 70\% & 90\% & 17\% & 90\%\\

    Ours & $\text{Wafer2Spike}^{2C}$ & 8:2 & 95\% & 97\% & \textbf{99\%} & \textbf{98\%} & 69\% & 96\% & 75\% & \textbf{99\%} & 47\% & \textbf{100\%}\\

    Ours & $\text{Wafer2Spike}^{3C}$ & 8:2 & 97\% & 99\% & \textbf{99\%} & 92\% & 83\% & 96\% & 84\% & 95\% & 53\% & \textbf{100\%}\\
    
    Ours & $\text{Wafer2Spike}^{4C}$ & 8:2 & \textbf{98\%} & 99\% & 98\% & 95\% & \textbf{87\%} & \textbf{98\%} & \textbf{87\%} & 95\% & \textbf{69\%} & \textbf{100\%}\\
    Ours & $\text{Wafer2Spike}^{4C}$ & 8:2 & \textbf{98\%} & $\prescript{F1}{}{99\%}$ & $\prescript{F1}{}{99\%}$ & $\prescript{F1}{}{94\%}$ & $\prescript{F1}{}{88\%}$ & $\prescript{F1}{}{99\%}$ & $\prescript{F1}{}{89\%}$ & $\prescript{F1}{}{95\%}$ & $\prescript{F1}{}{55\%}$ & $\prescript{F1}{}{77\%}$\\
    \hline    
    \cite{genssler2021brain} & MC32+MLP & 7:3 & 95\% & \textbf{99\%} & 87\% & 89\% & 71\% & 94\% & 50\% & 71\% & 12\% & 95\%\\
    
    Ours & $\text{Wafer2Spike}^{2C}$ & 7:3 & 95\% & 97\% & 96\% & 98\% & 67\% & 96\% & 59\% & 89\% & 51\% & 88\%\\

    Ours & $\text{Wafer2Spike}^{3C}$ & 7:3 & \textbf{97\%} & \textbf{99\%} & 98\% & 98\% & \textbf{85\%} & 97\% & 81\% & \textbf{93\%} & 46\% & \textbf{100\%}\\
    
    Ours & $\text{Wafer2Spike}^{4C}$ & 7:3 & \textbf{97\%} & \textbf{99\%} & \textbf{99\%} & \textbf{99\%} & 81\% & \textbf{98\%} & \textbf{84\%} & 92\% & \textbf{67\%} & 88\%\\
    \hline

    \cite{tsai2020light} & DMC1 & 6:1:3 & 97\% & 98\% & 93\% & 83\% & 84\% & \textbf{98\%} & 79\% & 83\% & \textbf{61\%} & 83\%\\
    
    Ours & $\text{Wafer2Spike}^{2C}$ & 6:1:3 & 95\% & 98\% & \textbf{98\%} & 96\% & 68\% & 96\% & 81\% & 93\% & 32\% & \textbf{100\%}\\

    Ours & $\text{Wafer2Spike}^{3C}$ & 6:1:3 & 96\% & 98\% & 97\% & 93\% & 84\% & \textbf{98\%} & 80\% & 88\% & 54\% & 93\%\\
    
    Ours & $\text{Wafer2Spike}^{4C}$ & 6:1:3 & \textbf{97\%} & \textbf{99\%} & 97\% & \textbf{97\%} & \textbf{87\%} & 96\% & \textbf{84\%} & \textbf{97\%} & 49\% & 86\%\\
    \hline
    
    \cite{liao2022wafer} & $P^2$-Net & 8:1:1 & 96\% & $\mathbf{\prescript{R}{}{99\%}}$ & $\prescript{R}{}{93\%}$ & $\prescript{R}{}{81\%}$ & $\prescript{R}{}{73\%}$ & $\mathbf{\prescript{R}{}{97\%}}$ & $\prescript{R}{}{60\%}$ & $\prescript{R}{}{91\%}$ & $\prescript{R}{}{29\%}$ & $\prescript{R}{}{90\%}$\\

    Ours & $\text{Wafer2Spike}^{2C}$ & 8:1:1 & 95\% & $\prescript{R}{}{98\%}$ & $\prescript{R}{}{98\%}$ & $\prescript{R}{}{96\%}$ & $\prescript{R}{}{70\%}$ & $\prescript{R}{}{96\%}$ & $\prescript{R}{}{72\%}$ & $\mathbf{\prescript{R}{}{98\%}}$ & $\prescript{R}{}{42\%}$ & $\mathbf{\prescript{R}{}{100\%}}$\\

    Ours & $\text{Wafer2Spike}^{3C}$ & 8:1:1 & \textbf{97\%} & $\mathbf{\prescript{R}{}{99\%}}$ & $\prescript{R}{}{98\%}$ & $\mathbf{\prescript{R}{}{100\%}}$ & $\prescript{R}{}{79\%}$ & $\mathbf{\prescript{R}{}{97\%}}$ & $\mathbf{\prescript{R}{}{86\%}}$ & $\prescript{R}{}{92\%}$ & $\mathbf{\prescript{R}{}{62\%}}$ & $\mathbf{\prescript{R}{}{100\%}}$\\
    
    Ours & $\text{Wafer2Spike}^{4C}$ & 8:1:1 & \textbf{97\%} & $\mathbf{\prescript{R}{}{99\%}}$ & $\mathbf{\prescript{R}{}{100\%}}$ & $\prescript{R}{}{96\%}$ & $\mathbf{\prescript{R}{}{85\%}}$ & $\prescript{R}{}{96\%}$ & $\mathbf{\prescript{R}{}{86\%}}$ & $\mathbf{\prescript{R}{}{98\%}}$ & $\prescript{R}{}{54\%}$ & $\mathbf{\prescript{R}{}{100\%}}$\\
    \hline
  \end{tabular}
  
  \begin{tablenotes}
      \small
      \item * Under the Split Ratio column, the value before the colon indicates the training set ratio, while the value after the colon represents the ratio of either the test set or a validation and test set. Samples are randomly selected to populate each set. 
      \item * \emph{\textbf{R}}: Recall, \emph{\textbf{F1}}: F1 score,  \emph{\textbf{H}}: The authors of \cite{alawieh2020wafer} used \cite{wu2014wafer}'s methodology with their own training set.
      \item * \emph{\textbf{2C}}, \emph{\textbf{3C}}, and \emph{\textbf{4C}} represent the number of spiking based convolutional layers in Wafer2Spike.
    \end{tablenotes}
  \endgroup
  \label{tab:main_results}
   \end{threeparttable}
   \vspace{-12pt}
\end{table*}

% \subsection{Evaluation Metrics}
% Accuracy and macro-average F1 score are used to evaluate the performance of Wafer2Spike. Accuracy is the ratio of correctly classified samples to the total number of samples, evaluating the overall effectiveness across all wafer map patterns. However, this metric can be skewed by imbalances in the dataset. Therefore, to assess performance on individual patterns, we use the macro-average F1 score. 
% \begin{align} \label{eq: metrics}
%     & \text{Accuracy} = \frac{\sum_{c=1}^C \text{TP}_c + \text{TN}_c}{\sum_{c=1}^C (\text{TP}_c + \text{TN}_c + \text{FP}_c + \text{FN}_c)}, \nonumber \\
%     & \text{Recall}_c = \frac{\text{TP}_c}{\text{TP}_c + \text{FN}_c}, \text{Precision}_c = \frac{\text{TP}_c}{\text{TP}_c + \text{FP}_c}, \; \mathrm{and}
%     \nonumber \\
%     & F1_{\text{macro}} = \frac{1}{C} \sum_{c=1}^C \Bigg(2 \times \frac{\text{Precision}_c \times \text{Recall}_c}{\text{Precision}_c + \text{Recall}_c}\Bigg), 
% \end{align}
% where $\text{TP}_c$, $\text{TN}_c$, $\text{FP}_c$, and $\text{FN}_c$ are true positive, true negative, false positive, and false negative rate, respectively, for class $c$.

\subsection{Main Results}
Table~\ref{tab:main_results} compares the performance of Wafer2Spike with other leading methods on the WM-811k dataset. It also includes an ablation study that varies the number of spiking-based convolutional layers in the architecture, resulting in models with two, three, or four layers. The two-layer configuration, $\text{Wafer2Spike}^{2C}$, achieves an accuracy of 95\% when the dataset is split in an 8:2 ratio, outperforming SVM and CNN-based approaches~\cite{wu2014wafer,alawieh2020wafer}. With a 7:3 split ratio, $\text{Wafer2Spike}^{2C}$ exceeds the performance reported by Genssler et al.~\cite{genssler2021brain} for all patterns except No-Pattern, Edge-Loc, and Near-Full. In a 6:1:3 data split scenario, $\text{Wafer2Spike}^{2C}$ achieves good accuracy while also adapting well to a smaller volume of training data. $\text{Wafer2Spike}^{3C}$ achieves an accuracy of 97\% across all split ratios, surpassing the two-layer model and all other competing methods~\cite{wu2014wafer,alawieh2020wafer,tsai2020light,genssler2021brain,liao2022wafer}. Finally, $\text{Wafer2Spike}^{4C}$, achieves 98\% accuracy, outperforming all preceding configurations and competing techniques, showing challenges only for No Pattern (8:2 split), and Edge-Ring and Scratch (6:1:3 split). For the 8:1:1 split, the only pattern it does not surpass is Edge-Ring. 

%There is a clear correlation between the number of spiking-based convolutional layers and performance improvements in Wafer2Spike. Each incremental layer significantly improves accuracy. Wafer2Spike also shows stable performance across different data splits, indicating that the model can reliably operate under different proportions of training and testing data, which is crucial for real-world applications where data distribution can vary. 

%In summary, $\text{Wafer2Spike}^{4C}$ establishes a new benchmark for accuracy and efficacy, achieving near-perfect accuracy rates and outperforming all previously mentioned methods. Our results demonstrate the effectiveness of deep spiking-based convolutional architectures in handling complex wafer map patterns in the semiconductor industry.

\subsection{Computational Efficiency}

The energy consumed during inference is modeled in terms of the number of floating point operations (FLOPs) and synaptic operations (SOPs) required. For an SNN, the energy required for a specific layer $L$ is calculated as $\text{Power}(L) = 77 \text{fJ} \times \text{SOPs}(L)$. Here, 77 fJ represents the energy consumed per SOP, based on empirical data~\cite{indiveri2015neuromorphic}. The number of synaptic operations within each layer $L$ is estimated as $\text{SOPs}(L) = T \times \gamma \times \text{FLOPs}(L)$,
where $T$ indicates the number of time steps needed for the simulation, $\gamma$ is the firing rate of the input spike train at layer $L$, 
and $\text{FLOPs}(L)$ represents the floating point operations estimated for that layer.  The firing rate is estimated as $\gamma = \frac{N_{Spk}}{T}$, where
$N_{Spk}$ represents the number of spikes emitted by the neuron during the observation period $T$. For DNN models, the energy consumption of the layer $L$ is estimated to be $\text{Power}(L) = 12.5 \text{pJ} \times \text{FLOPs}(L)$. The FLOPs performed within the network are calculated as a sum of 
\begin{align} \label{eq:flops}
    & {FLOPs}_{conv} = c^L \times d^L \times w_{c^L} \times h_{c^L} \times w_{w^L} \times h_{w^L} \times 2, \nonumber \\
    & {FLOPs}_{fc} =  u^L \times u^{L-1} \times 2,
\end{align}
where ${FLOPs}_{conv}$ denotes number of floating point operations for convolutional layers and ${FLOPs}_{fc}$, number of calculations for fully connected layers. The following parameters are used to calculate the computational requirements of each layer within the network: $c^L$, the number of output feature maps within layer $L$; $d^L$, the number of input channels; $w_{c^L}$ and $h_{c^L}$, the width and height of the output feature maps, respectively; $w_{w^L}$ and $h_{w^L}$, the width and height of the convolutional kernels, respectively; and $u^L$, the number of neuron units present in layer $L$. 

%${FLOPs}_{conv}$ accounts for the fact that each element of the output feature map at layer $L$ is produced by performing $w_{w^L} \times h_{w^L} \times d^L$ multiplications and $w_{w^L} \times h_{w^L} \times d^L - 1$ additions, but, typically, each multiply-accumulate operation (a multiplication followed by an addition) is counted as two operations and hence multiplied by two. For the fully connected layer, ${FLOPs}_{fc}$ assumes that each neuron requires $u^{L-1}$ multiplications and an equal number of additions; consequently, the total FLOPs for each neuron are doubled. 

\begin{table}[t!]
\small
\centering
\fontsize{6.5}{12}\selectfont
\caption{Theoretical estimation of energy consumption.}

\begingroup
\setlength{\tabcolsep}{6pt} % Default value: 6pt
\renewcommand{\arraystretch}{.90}
  \begin{tabular}{|cp{1.5cm}ccc|}
    \hline
    Reference & Model & Accuracy & FLOPs / SOPs ($10^9$) & Power (mJ) \\
    \hline
    \cite{alawieh2020wafer} & CNN & 94\% & 0.2391 & 2.9884 \\
    \cite{genssler2021brain} & MC32+MLP & 96\% & 0.0005 & 0.0062  \\
    \cite{tsai2020light} & DMC1 & 97\% & 0.4184 & 5.2299\\
    \cite{liao2022wafer} & $P^2$-Net & 96\% & 0.0090 & 0.1131 \\
    Ours & $\text{Wafer2Spike}^{2C}$ & 95\% & 3.1391 & 0.2417 \\
    Ours & $\text{Wafer2Spike}^{3C}$ & 97\% & 19.3960 & 1.4935 \\
    Ours & $\text{Wafer2Spike}^{4C}$ & 98\% & 21.9972 & 1.6938  \\
    \hline
  \end{tabular}

   % \begin{tablenotes}
   %    \small
   %    \item \hspace{2.1cm} * $1\text{ Joule (J)} = 10^3 \text{ millijoules (mJ)}= 10^{12}\text{ picojoules (pJ)}=10^{15}\text{ femtojoules (fJ)}$
   %    \item \hspace{2.1cm} * In our experiment, we have set T to 10
   %    \item \hspace{2.1cm} * \emph{\textbf{a}}, \emph{\textbf{b}}, \emph{\textbf{c}} represent with respect to $\text{Wafer2Spike}^{4C}$, $\text{Wafer2Spike}^{3C}$, $\text{Wafer2Spike}^{2C}$ respectively.
      
   % \end{tablenotes}
  
  \endgroup
  \label{tab:th_energy_results}
   \vspace{-12pt}
\end{table}

Table~\ref{tab:th_energy_results} presents the estimated energy consumption (in mJ) of the different architectures. Wafer2Spike performs better than DNNs; the basic model, $\text{Wafer2Spike}^{2C}$, is approximately 12.5x and 22x more efficient than CNN and DMC1, respectively, while the most complex model, $\text{Wafer2Spike}^{4C}$, improves efficiency by 1.75x and 3x. The MC32+MLP and $P^2$-Net models achieve the best energy efficiency. However, it is important to note that \eqref{eq:flops} only models energy consumption during the inference process but does not account for preprocessing steps such as generating hypervectors for MC32+MLP and extracting pixel-level details for $P^2$-Net.  

\section{Conclusion}\label{sec:conclusion}
This paper has developed Wafer2Spike, a spiking neural network for wafer pattern classification. The results obtained using the WM-811k benchmark confirm that Wafer2Spike outperforms existing methods, including DNN-based ones, in addition to being very computationally efficient. We have released Wafer2Spike as open source software available at \url{https://github.com/abhishekkumarm98/Wafer2Spike}.

\section{Acknowledgment}
This material is based on work supported by the National Science Foundation under Grant Number 2209745.

%\balance

% \bibliographystyle{IEEEtran}
% \bibliography{Wafer}
% \printbibliography %Prints bibliography

\end{document}